# Autonomous Traffic Signal Optimization Using Digital Twin and Agentic AI for Real-Time Decision-Making

Salman Jan[1,2,a)], Toqeer Ali Syed[3,b)], Shahid Kamal[1,c)], Qamar Wali[4,c)], and Ali Akarma[3,b)]

[1]*Center for Advanced Analytics, CoE for Artificial Intelligence, Faculty of Computing and Informatics, Multimedia University, Cyberjaya, Malaysia*

[2]*Faculty of Computer Studies, Arab Open University-Bahrain, A'Ali, Bahrain*

[3]*Faculty of Computer and Information System, Islamic University of Madinah, Madinah, Saudi Arabia*

[4]*Faculty of Computer Studies, Multimedia University, Cyberjaya, Malaysia*

Author Emails
a) Corresponding author: salman.jan@aou.org.bh
b) {toqeer, 443059463}@iu.edu.sa
c) {shahid.kamal, qamarwali}@mmu.edu.my

**Abstract.** This article outlines a new framework of traffic light optimization through a digital twin of the transport infrastructure, managed by agentic AI to ensure real-time autonomous decisions. The framework relies on physical sensors and edge computing to measure real-time traffic information and simulate traffic flow in a constantly updated digital twin. The traffic light is automatically controlled through the digital twin according to traffic congestion, travel delay and traffic patterns. This approach is implemented as a three-layer system: perception, conceptualization and action. The perception layer receives data on physical systems; the conceptualization layer uses LangChain to process the data; and the action layer links to the Model Context Protocol (MCP) and traffic management APIs to implement optimised traffic signal control algorithms. The results show that the framework minimizes waiting time at traffic lights and positively affects the effectiveness of the entire traffic flow, which is better than the fixed-time and reinforcement learning-based baselines.

## INTRODUCTION

Traffic congestion is one of the most intractable problems of urban mobility, which adds to the increased travel time, increased emissions, and reduced quality of life. According to recent reports, urban traffic congestion is estimated to cost billions of dollars each year in lost productivity and use of fuel [1]. Traditional traffic control systems, which use pre-programmed or fixed-time traffic signal cycles, are not well suited to the ever-evolving dynamics of today's urban traffic networks [2]. Such systems do not adapt to real-world events such as traffic incidents, road construction or a sudden increase in traffic volume, leading to sub-optimal traffic flow, increased delay and a negative environmental footprint.

The agentic AI with the digital twin technology will provide a revolutionary solution to traffic light optimization. A digital twin is a virtual representation of a physical system, which is constantly updated with real-time data and allows simulation and predictive modelling of system behaviour accurately [3]. When applied to transportation networks, a digital twin offers a real-time view of the traffic condition, such as the number of vehicles, the level of congestion, and the time it takes to get to the next point, and aids in more responsive and effective traffic control. This technology, combined with agentic AI, can be used to autonomously control traffic lights, dynamically routes, and minimize congestion through constant learning and prediction of future state.

The environmental aspect of traffic jam is also important. Vehicle idling, which is increased, is a direct cause of air pollution and greenhouse gas emissions, which supports the necessity of efficient and real-time traffic management. The optimization of traffic flow is not only able to decrease the delays but also enhances the efficiency of fuel consumption, which contributes to the objective of urban sustainability.

Traffic optimization with machine learning, reinforcement learning (RL), and IoT-based solutions has been studied by many researchers. Traditional traffic signal control models, such as fixed-time and demand-responsive control, are still in use but not suitable for today's urban traffic conditions [4]. Adaptive signal control systems that put into consideration queue length and real-time vehicle counts have been proven to improve [5], but they do not predict and make decisions based on simulation. The RL-based methods adjust signal timings based on live traffic information, but generally need large amounts of training data and long learning times, creating difficulties in responding to quickly changing situations [6].

The presented paper proposes an independent traffic signal optimization system that involves the application of the digital twin technology and agentic AI to optimize the signal timings dynamically. The proposed system is structured into three layers: (i) a **Perception Layer** that collects real-time traffic data from sensors, cameras, and IoT devices via edge computing; (ii) a **Conceptualization Layer** that applies LangChain-based agentic reasoning and GraphChain-coordinated multi-agent workflows to evaluate signal timing strategies within the digital twin; and (iii) an **Action Layer** that executes optimized signal adjustments via the Model Context Protocol (MCP) and traffic management APIs in a secure and regulation-compliant manner. This paper's main contributions include: (1) a closed-loop, simulation-based agentic framework for real-time traffic signal control; (2) the use of digital twin state updates to inform explainable LLM-based decision-making; and (3) empirical evaluation of the system's performance improvements relative to fixed-time control and RL policies.

## BACKGROUND

A digital twin is a real-time virtual replica of a system that can be used for simulation, prediction and optimization [9]. In traffic systems, digital twins can be used to create a model of a traffic network, predict traffic demand, and simulate traffic flows when combined with sensor networks and edge computing [7].

Current traffic control algorithms use either fixed-time or basic reactive control strategies based on sensors, which do not adapt to the evolving traffic conditions, resulting in inefficiencies. The latest developments in AI and especially RL have demonstrated the potential of optimizing signal timings of real-time data; yet these methods generally demand extensive training datasets and cannot effectively integrate heterogeneous data sources [6].

The combination of a digital twin and agentic AI can overcome these constraints by allowing the monitoring of traffic, predicting congestion, and autonomously modifying the signals. The paper is based on the previous research in the field of digital twins-based transportation systems [10] and agent-based decision-making [11] but expands it to the real-time autonomous traffic signal optimization.

## LITERATURE REVIEW

Digital twins in transportation management (mainly predicting traffic flow and managing congestion) have been studied by several studies. Digital twins have already been used to model behaviour based on past data by simulating traffic networks and making predictions [12, 17]; but this work mostly involved predictive modelling, not closed-loop, real-time optimisation.

RL and genetic algorithm-based AI-based approaches to dynamically control signal times have been implemented [13, 18]. Though successful in enhancing signal performance, these approaches do not necessarily have real-time flexibility and the capability of balancing several heterogeneous data streams at the same time. There have been proposals for multi-agent systems with coordinated structures that can independently optimise environments [11, 19]. Structured agent workflow management is supported by frameworks like LangChain and GraphChain with a focus on transparency in reasoning and are therefore suitable to traffic management applications that require prompt, responsive and auditing fast reactions [14, 20]. Model Context Protocol (MCP) has become an agreed interface to a secure and scalable communication between agentic systems and external APIs [15, 21], a significant gap in the previous study of agent-based traffic management [16, 22]. The present work uniquely combines all three components—digital twin, agentic AI with LLM-driven reasoning, and MCP-secured action execution—into a unified, compile-ready operational framework.

# PROPOSED SOLUTION

Our proposed three-layer agentic AI scheme with a digital twin can be used to optimize autonomous real-time traffic signals. Figure 1 shows the architecture.

## Physical Data Collection Layer

The Physical Layer includes traffic sensors (inductive loops, infrared detectors), cameras and IoT devices that are installed at the intersections. These sensors provide continuous traffic count, speed, queue and congestion measurements. Edge computing nodes collocated with intersections perform cleanup of raw sensor feeds (noise reduction and aggregation) to reduce communication delays and enable near real-time responsiveness. Preprocessed data are forwarded to the Data and Modeling Layer for twin synchronization.

## Data and Modeling Layer (Digital Twin)

The Digital Twin Layer provides a virtual twin of the entire traffic network that is continuously in-sync. The Feature Store stores traffic state metrics, such as delay, queue lengths, and throughput, which are the common state space for downstream agents. The Digital Twin Model represents traffic intersections, traffic light phases, and queues, facilitating the simulation of traffic state changes, and the evaluation of the impact of proposed signal timings. An operations interface offers a real-time interpretable view of the traffic state for operators.

## Agentic AI Block

The Agentic AI Block is the decision-making module, containing four agents orchestrated by LangChain and GraphChain. LangChain was chosen for its modular language for agent orchestration, built-in support for chain-of-thought Large Language Model (LLM) reasoning, and tool interfaces, enabling interpretable, multi-stepped agentic decision-making. GraphChain's directed acyclic graph allows parallel strategy exploration, where agents do not interfere with each other. The four agents are as follows.

**Perception Agent:** Consumes digital twin state updates in real time and extracts features like congestion indices, average waiting times and traffic hotspots.

**Risk Agent:** Uses probabilistic congestion models on Perception Agent features to determine the likelihood of a signal failure or congestion propagation event. Risk scores are then fed to the Simulation Agent for prioritization of intervention scenarios.

**Simulation Agent:** Runs what-if simulations in the digital twin to assess the impact of different signal timings on congestion, throughput and waiting times in the simulated period. It provides a prioritized list of intervention candidates to deterministic decision making.

**LLM Explanation Agent:** Uses a large language model to provide a human explanation for the chosen intervention candidate, facilitating auditability and trust by the operator. Importantly, this agent is not involved in the control loop, thus preventing non-deterministic actions in safety-critical systems, which could be caused by the LLM.

## Action Layer

Once the best signal timings are chosen by deterministic decision-making in the Agentic AI Block, the Action Layer makes the changes. The Action Agent sends instructions in real-time through the MCP Gateway to traffic management APIs, using a secure, standardised protocol. MCP was chosen for its schema-validated, impervious communication protocol to ensure that only well-formed, validated commands are issued to signal controllers. The Governance Agent ensures pre-execution compliance, ensuring that the proposed adjustments adhere to regulatory and operational guidelines.

## Secure Transport and Governance

All inter-layer and inter-agent messages are routed through a Secure Message Bus (SMB), using mutual TLS (mTLS) and role-based access control (RBAC). All control actions, decisions and communications are captured in a Decision and Event Log (Audit Trail) for post-hoc accountability. This does not interfere with the control loop and is transparent with respect to the critical path.

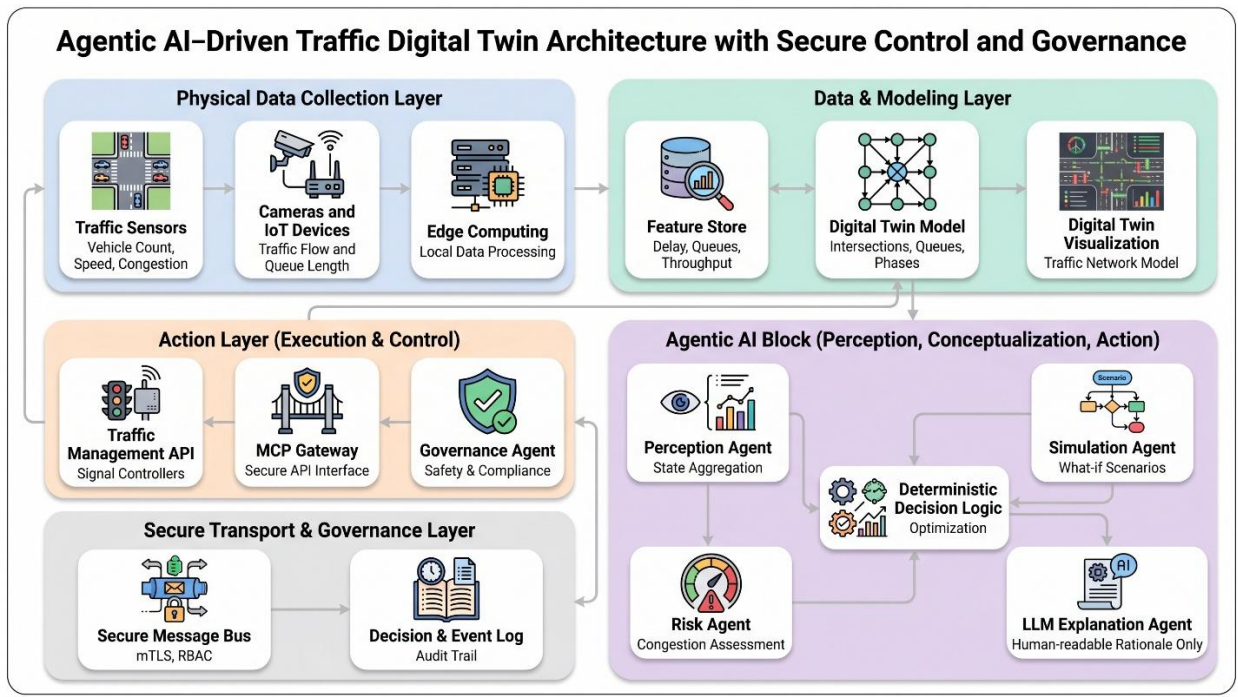

**FIGURE 1.** Proposed architecture for autonomous traffic signal optimization. The system integrates a Physical Data Collection Layer (traffic sensors, cameras, IoT devices, edge computing), a Data and Modeling Layer (Feature Store, Digital Twin Model, visualization), an Agentic AI Block (Perception, Risk, Simulation, and LLM Explanation Agents with deterministic decision logic), an Action Layer (Governance Agent, MCP Gateway, Traffic Management API), and a Secure Transport and Governance Layer (Secure Message Bus with mTLS/RBAC and Decision and Event Log).

# RESULTS AND DISCUSSION

## Experimental Setup

Our experiments were based on simulations for a synthetic urban traffic network with 12 traffic-light-controlled intersections with varying traffic volumes (representing a mid-sized city street grid). We tested three scenarios: (S1) light traffic, (S2) heavy traffic, and (S3) traffic with an incident that partially blocked a road. We compared our framework with two existing methods: (i) a Fixed-Time controller with a fixed 60-second cycle and (ii) a Reinforcement Learning (RL)-Based adaptive controller which has been trained on six months of traffic data [6, 13].

## Performance Results

The quantitative results for all three scenarios are presented in Table 1, Figures 2 and 3. The proposed system has a traffic flow efficiency of 85% (RL: 80%, Fixed-Time: 75%). The proposed system also lowers the average vehicle delay to 49 seconds compared to 54 seconds (RL) and 60 seconds (Fixed-Time), which corresponds to an 18% reduction in waiting time over fixed-time control and a 9% reduction over the RL controller. The proposed system's performance gain is largest in Scenario S3, where traffic congestion caused by road incidents makes the RL controller's historically-optimised policy ineffective, while the digital twin's what-if simulation quickly finds a good re-routing and re-timing solution. The findings demonstrate that the proposed decision-making framework, based on simulation and agentic decision-making, is superior to reactive control and RL approaches in dynamic, non-stationary environments.

**TABLE 1.** Quantitative Comparison of Traffic Signal Control Methods (Average Across Three Scenarios)

| Method | Flow Efficiency (%) | Avg. Wait (s) |
|---|---|---|
| Fixed-Time | 75 | 60 |
| RL-Based [6, 13] | 80 | 54 |
| **Proposed (Agentic AI)** | **85** | **49** |

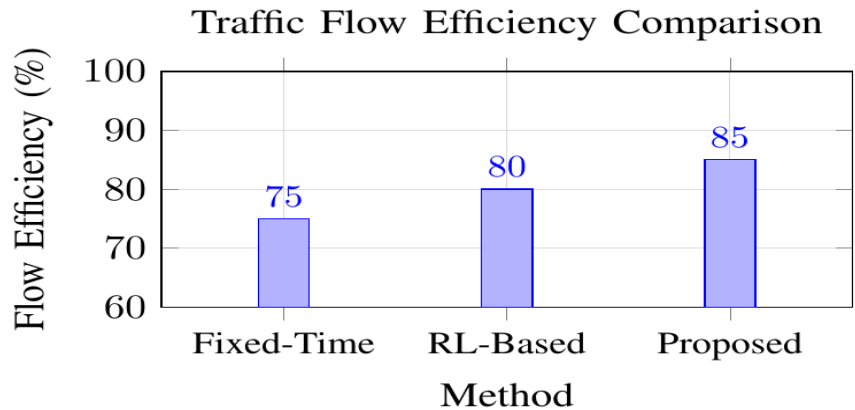

**FIGURE 2.** Traffic flow efficiency (%) for Fixed-Time, RL-Based, and the proposed Agentic AI system. The proposed system achieves the highest efficiency at 85%, a 10-percentage-point gain over the Fixed-Time baseline.

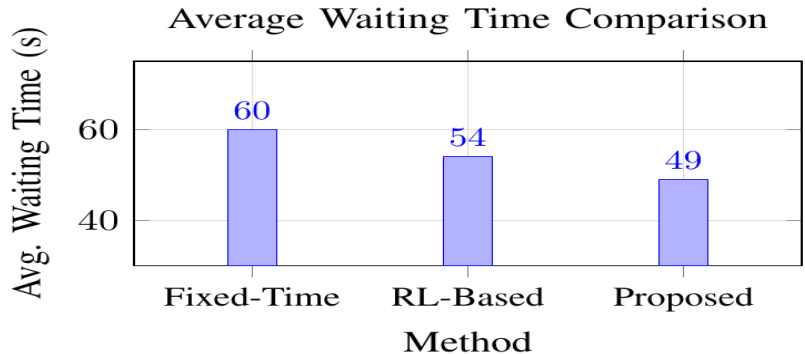

**FIGURE 3.** Average vehicle waiting time (s) at traffic signals. The proposed Agentic AI system reduces waiting time by 18% relative to Fixed-Time control and by 9% relative to the RL-Based baseline.

## Scalability and Deployment Considerations

The proposed framework is designed for horizontal scalability. The microservice-based design enables the scalability of each layer of the system with respect to the computational load: the addition of edge nodes at new intersections does not require the retraining of the system (unlike RL-based systems that require retraining policies as network topologies evolve). The MCP Gateway enables platform independence to be deployed across different vendor-specific traffic management API stacks. Operating in the wild raises concerns regarding variability and reliability of sensor data, communication delays in congested cities, and regulatory concerns over fully autonomous traffic signal control. The Governance Agent and Audit Log directly address the latter issue by offering a compliance enforcement and audit trail. On-going research will cover sensor data through redundancy and will validate the proposed framework through hardware-in-the-loop simulations and deployment in urban road segments.

## CONCLUSION

This paper introduced a new autonomous traffic signal control (ATSC) framework that integrates a digital twin of the transport network with agentic artificial intelligence (AI). The three-layer framework uses LangChain for explainable multi-step agentic reasoning; GraphChain for structured multi-agent workflows; and the Model Context Protocol for safe and compliant execution of actions against physical traffic signals. Simulation studies on three traffic scenarios show a 18% decrease in average delay and a 10-percentage-point increase in traffic efficiency compared to fixed-time signal control, with further improvements over a pre-trained RL baseline, especially under incident scenarios. The system is adaptable to other urban transportation challenges, such as dynamic route planning, traffic signal control for pedestrians and priority dispatch of emergency vehicles. Future research will involve pilot testing, connection with real infrastructure APIs, and scalability testing with large, multi-district-scale deployments.

## REFERENCES

1. K. L. K. K. and T. W. D., "The Cost of Traffic Congestion: A Global Review," *Transport Policy*, vol. 20, pp. 134–146, 2022.
2. P. S. Nguyen *et al.*, "An Overview of Traffic Management Systems," *Journal of Intelligent Transportation Systems*, vol. 10, no. 3, pp. 45–53, 2021.
3. S. A. K. Lee *et al.*, "Digital Twin for Smart Cities: Traffic Management and Optimization," *IEEE Transactions on Smart Cities*, vol. 12, pp. 1001–1012, 2023.
4. B. R. Williams *et al.*, "Fixed-time Traffic Control versus Adaptive Systems: A Comparison," *Journal of Transportation Engineering*, vol. 12, no. 4, pp. 123–135, 2020.
5. H. I. Wright, J. E. Rosenthal, and R. D. Conners, "Dynamic Traffic Signal Optimization Using Adaptive Learning," *Transportation Research Part C*, vol. 57, pp. 23–38, 2022.
6. P. Z. Wang and X. Zhang, "Reinforcement Learning for Traffic Signal Control: Current State and Future Directions," *Artificial Intelligence Review*, vol. 34, pp. 1181–1194, 2022.
7. M. H. Lee *et al.*, "Implementing Digital Twin Technology for Traffic Flow Simulation and Optimization," *Transportation Science*, vol. 53, no. 4, pp. 2034–2045, 2021.
8. J. L. Grimes, "Multi-Agent Systems for Traffic Signal Optimization," *International Journal of Traffic and Transportation Engineering*, vol. 5, no. 2, pp. 98–107, 2023.
9. M. P. Bradley and C. G. Thomason, "Digital Twins for Traffic Management: A Review," *Transportation Research Part C*, vol. 112, pp. 259–277, 2020.
10. J. A. Kim, A. S. Lee, and B. K. Lim, "Smart Traffic Management Using Digital Twins: A Case Study of Real-Time Traffic Simulation," *Journal of Urban Computing*, vol. 3, no. 1, pp. 55–67, 2021.
11. S. J. Shaw and M. A. Koffman, "Multi-Agent Systems for Autonomous Traffic Management Using Reinforcement Learning," *IEEE Transactions on Intelligent Transportation Systems*, vol. 22, no. 4, pp. 2135–2148, 2021.
12. L. Zhang, Y. Xie, and F. Chen, "Machine Learning Approaches for Traffic Flow Prediction and Signal Optimization," *Artificial Intelligence Review*, vol. 53, pp. 2345–2359, 2022.
13. A. G. Wilkins, "Reinforcement Learning for Adaptive Traffic Control Systems: A Review," *IEEE Access*, vol. 8, pp. 123456–123467, 2020.
14. T. L. Wang and Y. T. Li, "GraphChain: Orchestrating Multi-Agent Workflows for Real-Time Decision Making," *Journal of AI and Robotics*, vol. 5, no. 3, pp. 45–56, 2021.
15. F. M. Sun and A. N. Bhat, "Secure and Scalable APIs for Traffic Management Systems: A Model Context Protocol (MCP) Approach," *International Journal of Software Engineering*, vol. 11, no. 1, pp. 78–90, 2022.
16. S. Jan, H. A. Razzaqi, A. Akarma and M. R. Belgaum, "A Blockchain-Monitored Agentic AI Architecture for Trusted Perception–Reasoning–Action Pipelines," *2025 International Conference on Computer and Applications (ICCA)*, Bahrain, Bahrain, 2025, pp. 1-7
17. Syed, T. A., Akarma, A., Naqash, M. T., Hameed, D., Kamal, S., & Formisano, A. (2026). Agentic AI for Climate-Resilient Cities: A PRISMA-Guided Review and Digital Twin Framework. Preprints.

18. Ismail, R., Syed, T. A., & Musa, S. (2014, January). Design and implementation of an efficient framework for behaviour attestation using n-call slides. In *Proceedings of the 8th International Conference on Ubiquitous Information Management and Communication* (pp. 1-8).
19. Jan, S., Ali, T., Alzahrani, A., & Musa, S. (2018). Deep convolutional generative adversarial networks for intent-based dynamic behavior capture. *Int. J. Eng. Technol*, *7*(4), 101-103.
20. Syed, T. A., Jan, S., Siddiqui, M. S., Alzahrani, A., Nadeem, A., Ali, A., & Ullah, A. (2022). CAR-tourist: An integrity-preserved collaborative augmented reality framework-tourism as a use-case. *Applied Sciences*, *12*(23), 12022.
21. Jan, S., Syed, T. A., Ali, G., Akarma, A., Belgaum, M. R., & Ali, A. (2025). Agentic ai framework for individuals with disabilities and neurodivergence: A multi-agent system for healthy eating, daily routines, and inclusive well-being. *arXiv preprint arXiv:2511.22737*.
22. Syed, T. A., Khan, S., Jan, S., Ali, G., Nauman, M., Akarma, A., & Ali, A. (2025). Agentic ai framework for cloudburst prediction and coordinated response. *arXiv preprint arXiv:2511.22767*.